%% file: acl_latex.tex
\documentclass[11pt]{article}

\usepackage[final]{acl}

\usepackage{times}
\usepackage{latexsym}

\usepackage[T1]{fontenc}
\usepackage[utf8]{inputenc}

\usepackage{microtype}
\usepackage{inconsolata}

\usepackage{amsmath}
\usepackage{amsfonts}
\usepackage{amssymb}

\usepackage{booktabs}
\usepackage{multirow}
\usepackage{array}
\usepackage{tabularx}
\usepackage{adjustbox}

\usepackage{algorithm}
\usepackage{algpseudocode}

\usepackage{graphicx}
\usepackage{xcolor}

\usepackage{caption}
\usepackage{subcaption}

\usepackage{mdframed}

\usepackage{enumitem}

\usepackage{hyperref}

\newcommand{\mysection}[1]{\vspace{2pt}\noindent\textbf{#1}}

\title{SGA: Plug \& Play Geometric Verification for Educational Video Synthesis}

\author{
Jhon Lopez\textsuperscript{1,2}\thanks{Work done during an internship at KAUST.} \hspace{0.5em}
Carlos Hinojosa\textsuperscript{2} \hspace{0.5em}
Bernard Ghanem\textsuperscript{2} \\[0.2em]
\textsuperscript{1}Universidad Industrial de Santander \\
\textsuperscript{2}King Abdullah University of Science and Technology\\
{\normalsize \url{https://jhedlp.github.io/projects/sga/}}
}

\begin{document}
\maketitle

\input{sec/0_abstract}

\input{sec/1_introduction}
\input{sec/2_related_works}
\input{sec/3_method}
\input{sec/4_experiments}
\input{sec/5_conclusions}

\input{sec/6_ethics}


\bibliography{custom}


% \appendix

% \section{Example Appendix}
% \label{sec:appendix}

\newpage
\input{sec/X_supp}

\end{document}

%% file: sec/0_abstract.tex
\begin{abstract}
Recent work leverages Large Language Models (LLMs) to generate executable code for pedagogical animations using libraries such as Manim. However, ensuring spatial correctness and visual legibility remains challenging, as existing frameworks emphasize pedagogical content while overlooking geometric occlusions. We propose the Symbolic Geometric Agent (SGA), a plug-and-play module for code-centric animation pipelines that intercepts LLM-generated code, performs partial execution to extract symbolic scene graphs, and applies targeted refinement when spatial conflicts are detected. We further introduce the Manim Visual Quality Score (MVQS), a deterministic rendering-free proxy for spatial integrity. Experiments on the MMMC-Code benchmark across four LLM backbones and two agentic pipelines show that SGA achieves a peak MVQS of 73.11 (Code2Video + GPT-5.1), corresponding to a 16.1\% relative improvement over the raw baseline, and improves MVQS in 7 of 8 backbone~$\times$~pipeline configurations. Additionally, we conduct a human evaluation showing that these improvements translate into human preference, with raters preferring SGA over the raw baseline in 84.4\% of comparisons and over a VLM-based critic in 65.0\%.
\end{abstract}

%% file: sec/1_introduction.tex
\section{Introduction}
\label{sec:introduction}

\input{figures/teaser}

Video has become a dominant medium for instructional content in modern digital education. In mathematics and science, structured animations require precise spatial organization, clear visual hierarchy, and coherent temporal transitions to effectively communicate abstract concepts~\cite{khalil2024video, navarrete2025closer, clark2023learning, mayer2024past}. As demand for scalable educational content grows, research increasingly focuses on automating the production of explanatory videos.

Recent advances in Large Language Models (LLMs) have enabled language-to-program generation, allowing natural language prompts to be translated into executable programs for visual content creation. Moving beyond direct synthesis in pixel space, emerging code-centric paradigms generate structured animation programs, such as Python scripts for the Manim engine~\cite{manimcommunity}, which are then rendered into educational videos~\cite{chen2025code2video}. This approach offers substantial advantages in controllability, reproducibility, and semantic transparency compared to diffusion-based or pixel-level video generation~\cite{lin2025creativity,shah2025challenge,huang2026genmac}. Representative systems such as Code2Video~\cite{chen2025code2video} and TheoremExplainAgent~\cite{ku2025theoremexplainagent} demonstrate that LLM-driven pipelines can produce long-form, coherent educational animations through multi-agent planning and refinement. Figure~\ref{fig:teaser_comparison} illustrates the limitations of current approaches and the motivation for our framework. 

Despite these advances, ensuring spatial correctness and visual legibility remains challenging. Educational animations require strict geometric consistency, in which text does not overlap, equations remain aligned, and objects preserve their scale relationships to maintain readability. Empirical analyses of state-of-the-art code-centric systems reveal that spatial layout remains the weakest component in automated generation. For instance, TheoremExplainAgent reports that element layout receives the lowest evaluation scores among all assessed criteria, even when factual accuracy and logical coherence are high~\cite{ku2025theoremexplainagent}. These issues are not merely aesthetic; in educational contexts, poor spatial organization increases cognitive load and can impair knowledge transfer \cite{clark2023learning, leiker2023prototyping}.

Current approaches primarily rely on post-hoc perceptual evaluation using Vision-Language Models (VLMs) to critique rendered video frames~\cite{chen2025code2video}. While effective for high-level assessment, VLM-based critique operates in pixel space, requires expensive full-video rendering, and produces coarse, non-deterministic feedback. Crucially, such evaluation lacks deterministic spatial verification. This reveals a methodological mismatch: when visual content is generated from structured code, verification should operate in the same symbolic domain. To address these limitations, we introduce the Symbolic Geometric Agent (SGA), a plug-and-play symbolic verification framework for code-centric educational video generation. We further introduce the Manim Visual Quality Score (MVQS), a deterministic rendering-free metric for spatial integrity. In summary, our contributions are threefold:

\begin{enumerate}
% \item We introduce the Symbolic Geometric Agent (SGA), a plug-and-play verification layer for code-centric animation frameworks that intercepts LLM-generated code to detect spatial conflicts and enable targeted refinement without modifying the generation pipeline.
\item We introduce the Symbolic Geometric Agent (SGA), a plug-and-play verification layer for code-centric animation frameworks that constructs symbolic scene graphs and computes geometric descriptors, such as axis-aligned bounding boxes and object distances, enabling deterministic spatial conflict detection and targeted refinement without modifying the generation pipeline.

\item We introduce the Manim Visual Quality Score (MVQS), a deterministic metric that evaluates visual layout quality without video rendering or VLM-based critique, focusing primarily on geometric legibility and spatial integrity rather than visual or educational quality.

% \item We propose a spatial verification approach based on partial execution of animation programs. This process constructs a symbolic scene graph and computes geometric descriptors such as axis-aligned bounding boxes and object distances, enabling deterministic conflict detection and code-level corrections.

\item We conduct extensive experiments on the MMMC-Code benchmark across four LLM backbones and two agentic pipelines, showing that SGA consistently improves spatial layout quality and outperforms VLM-based critique in spatial layout evaluation.

% \item We introduce the Manim Visual Quality Score (MVQS), a deterministic metric that evaluates visual layout quality without rendering or VLM-based critique. Experiments on the MMMC-Code benchmark show that SGA achieves a peak MVQS of 73.11 (Code2Video + GPT-5.1), a 16.1\% relative gain over the corresponding raw baseline, and improves MVQS in 7 of 8 backbone\,$\times$\,pipeline configurations. 

% Across all four backbones on Code2Video, SGA improves the spatial sub-score $S_\text{sp}$ by a mean of $+5.0$ points and the educational sub-score $S_\text{ed}$ by a mean of $+7.2$ points over the raw baseline (Supplemental Table~2), with peak gains of $+9.9$ on $S_\text{sp}$ and $+15.3$ on $S_\text{ed}$ for GPT-5.1.
\end{enumerate}

%% file: figures/teaser.tex
\begin{figure*}
    \centering
    \includegraphics[width=0.99\linewidth]{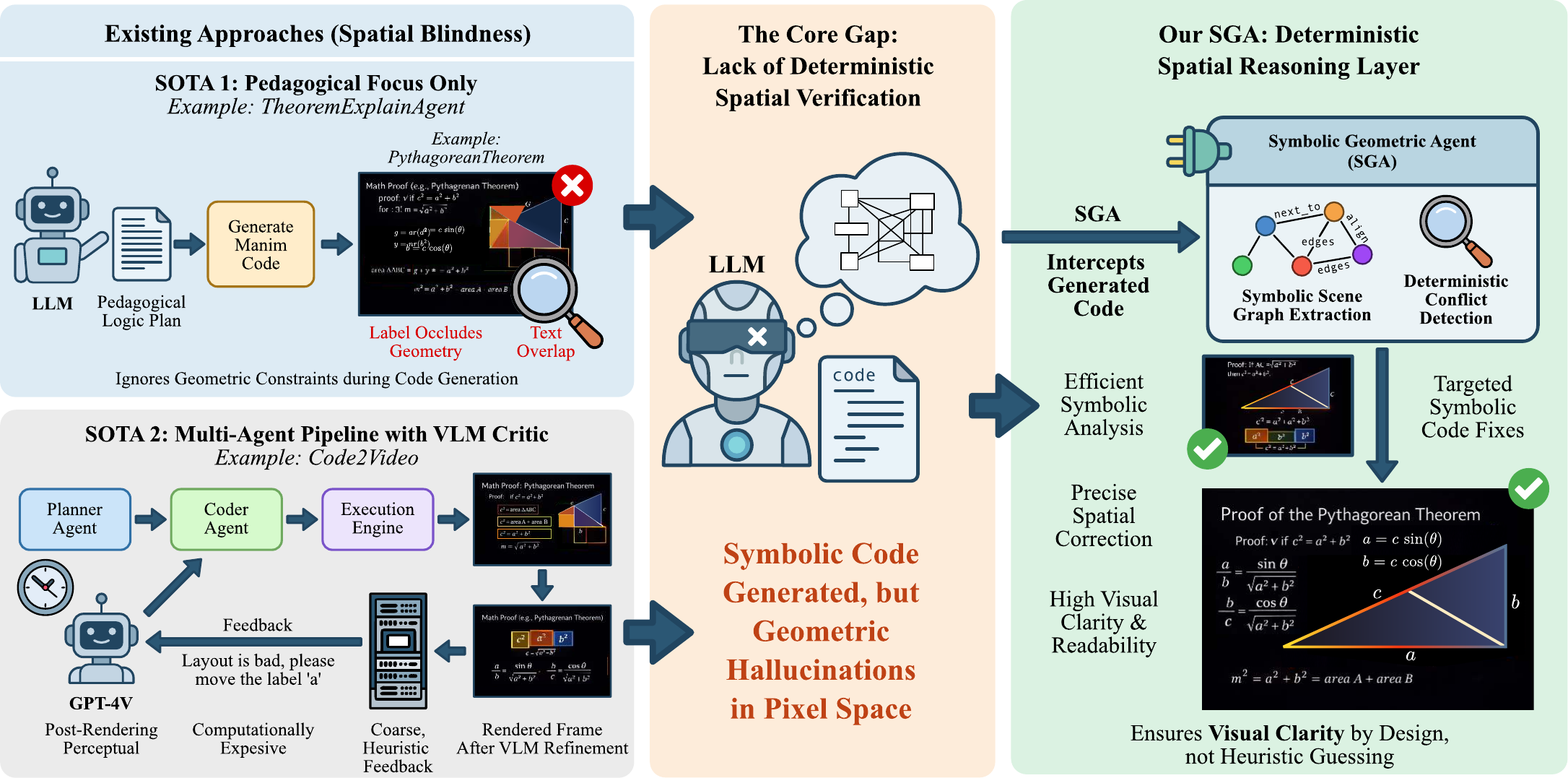}
    \caption{
    Current SOTA agents produce syntactically valid but visually occluded animations. (Left) \textit{TheoremExplainAgent} \cite{ku2025theoremexplainagent} neglects spatial constraints, while \textit{Code2Video} \cite{chen2025code2video} relies on stochastic, computationally expensive VLM feedback that fails to resolve fine-grained clashes. (Center) This lack of deterministic grounding induces ``geometric hallucinations'' in pixel space. (Right) Our SGA enables symbolic scene graph extraction and targeted refinement, ensuring spatial integrity by design prior to rendering.}
    \label{fig:teaser_comparison}
    \vspace{-1.25em}
\end{figure*}

%% file: sec/2_related_works.tex
\section{Related Work}
\label{sec:related_work}

\mysection{Code-Centric Educational Video Generation.} Traditional generative video research has largely focused on end-to-end models operating in pixel or latent spaces, including diffusion-based~\cite{gupta2024photorealistic, ho2022imagen, xing2024survey,lian2023llm} and autoregressive approaches~\cite{deng2024autoregressive, khachatryan2023text2video}. However, such models often struggle with the long-form reasoning, controllability, and spatial consistency required for instructional content. Recent advances in large language models have enabled code-centric generation paradigms in which LLMs produce executable animation programs through multi-step reasoning and iterative refinement. Frameworks such as ReAct~\cite{yao2022react} demonstrate how language models can interleave reasoning and external actions~\cite{shinn2023reflexion,madaan2023self,hong2024metagpt,vera2025multimodal,contreras2026projguard}, while specialized systems such as Code2Video~\cite{chen2025code2video} and TheoremExplainAgent~\cite{ku2025theoremexplainagent} show that structured program generation enables controllable and semantically coherent educational animations. Despite these advances, current systems primarily rely on internal LLM reasoning and lack mechanisms to verify the geometric consistency of generated layouts before rendering. Our work addresses this limitation through symbolic geometric verification directly in the code generation pipeline of systems such as Code2Video and TheoremExplainAgent.

\mysection{Automated Evaluation of Visual Quality.} Evaluation metrics for generative visual models have traditionally focused on pixel-level similarity or embedding-based alignment, often using measures such as CLIP~\cite{radford2021learning}. More recent systems employ Vision-Language Models (VLMs) as automated evaluators for generated images and videos~\cite{chen2025code2video,lee2024prometheus,chen2024mllm,liu2025your,hinojosa2026saves}, providing high-level assessments of visual quality and instruction alignment. However, VLM-based evaluation operates in pixel space and does not directly capture geometric violations in structured animation layouts. As a result, these approaches often provide coarse feedback that is difficult to translate into concrete code corrections. Our proposed MVQS metric instead evaluates layout quality using symbolic geometric descriptors derived from the animation scene representation, enabling feedback for refinement.

%% file: sec/3_method.tex
\section{Proposed Method}
\label{sec:method}

\begin{figure*}[t]
    \centering
    \includegraphics[width=0.95\textwidth]{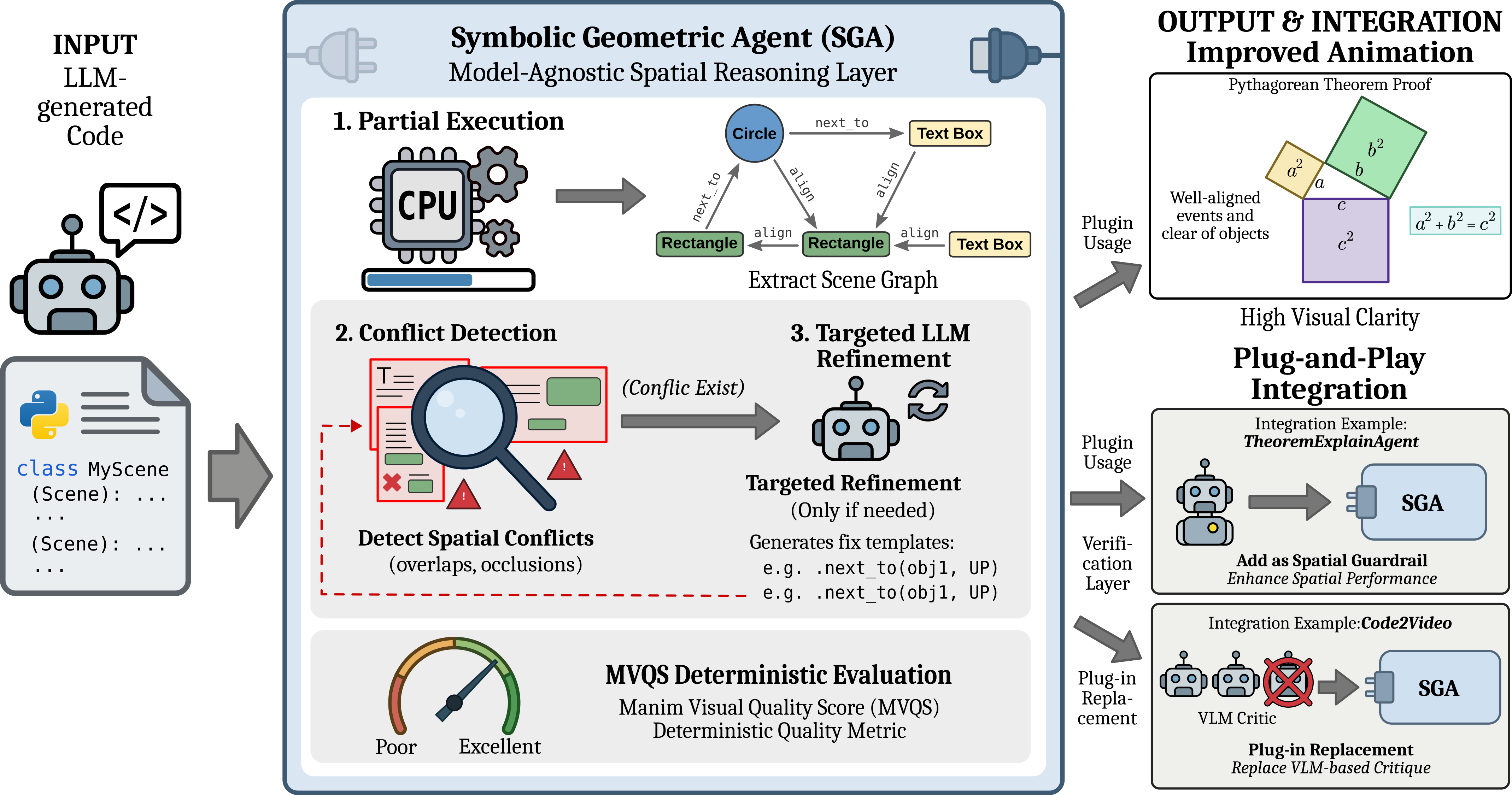}
    \caption{
    The SGA acts as a neuro-symbolic interceptor for spatial-linguistic grounding. (Left) The pipeline captures LLM-generated code pre-rendering. (Center) The SGA engine extracts a symbolic scene graph via partial execution, identifying spatial conflicts through deterministic AABB analysis. Violations prompt template-based refinement, quantified by the Manim Visual Quality Score (MVQS). (Right) SGA serves as a plug-and-play spatial guardrail, replacing stochastic VLM critics in frameworks like Code2Video and TheoremExplainAgent.}
    \label{fig:pipeline}
    \vspace{-1.25em}
\end{figure*}

Figure~\ref{fig:pipeline} illustrates our proposed Symbolic Geometric Agent (SGA), which mitigates spatial blindness in code-generation pipelines for educational video generation. Unlike VLM-based critics that rely on computationally expensive pixel-level video rendering and stochastic inference, SGA performs deterministic geometric verification directly on symbolic scene representations extracted via partial execution. Its modular, framework-agnostic design enables it to serve as a plug-and-play verification layer across generation frameworks.

% ------------------------------------------------------------------
\subsection{Plug-and-Play Architecture and Interception Flow}
\label{sec:overview}
% ------------------------------------------------------------------

The SGA operates as a non-invasive verification layer, treating animation synthesis as an iterative code-space optimization problem. Given a target topic $T$, a generative model $\mathcal{L}$ produces an initial script $C^{(0)}$. The SGA intercepts the execution flow and executes a four-stage refinement loop at each iteration $i$: (1) Symbolic Extraction, where $C^{(i)}$ undergoes low-latency partial execution to derive a symbolic scene graph $\mathcal{S}^{(i)}$, bypassing full video rendering overhead (Sec.~\ref{sec:extraction}); (2) Deterministic Conflict Detection, wherein $\mathcal{S}^{(i)}$ is evaluated for spatial violations via exact Axis-Aligned Bounding Box (AABB) analysis, replacing stochastic VLM heuristics with reproducible geometric verification (Sec.~\ref{sec:overlap}); (3) Actionable Feedback, involving a compiler $\mathcal{F}$ that translates violations into a structured diagnostic report $F^{(i)}$ containing pre-computed fix vectors and canonical Manim correction templates (Sec.~\ref{sec:feedback}); and (4) Targeted Refinement, where the LLM performs precise, line-level edits conditioned on the symbolic feedback:
\begin{equation}
C^{(i+1)} = \mathcal{L}(T, C^{(i)}, F^{(i)})
\end{equation}
Refinement terminates after a fixed budget of $n$ iterations, effectively bounding computational overhead while allowing the LLM to iteratively incorporate symbolic grounding constraints. A summarized version of the algorithm is provided in the supplemental material (Sec. 6).

% ------------------------------------------------------------------
% ------------------------------------------------------------------
\subsection{Symbolic Scene Graph Extraction}
\label{sec:extraction}
% ------------------------------------------------------------------

SGA avoids high-latency rendering by repurposing the Manim runtime as a symbolic execution engine. The system parses source code into an Abstract Syntax Tree (AST) and instrumentally rewrites the \texttt{construct()} method to wrap each statement in a granular \texttt{try-except} block. This fault-tolerant execution isolates runtime exceptions, such as invalid attribute calls common in LLM outputs, allowing the agent to skip faulty instructions and extract spatial primitives from the remaining valid code. 

\paragraph{Geometric Formalization.} For each Manim object (\texttt{mobject}) $m_j$ extracted into the set of all scene objects $\mathcal{M}$, the SGA derives a deterministic record $\mathcal{R}_j = \langle \phi_j, \mathbf{c}_j, \text{bb}_j, \psi_j, \ell_j \rangle$. (1) Semantic label $\phi_j \in \Phi$ is assigned from a four class taxonomy $\Phi = \{\textsc{Text, Geom, Mark, Deco}\}$\cite{ericson2004real}. (2) Spatial geometry is defined by centroid $\mathbf{c}_j = (c^x_j, c^y_j)$ and dimensions $(w_j, h_j)$, yielding Axis Aligned Bounding Box: $x^{\min/\max}_j = c^x_j \pm w_j/2$ and $y^{\min/\max}_j = c^y_j \pm h_j/2$. (3) Source grounding $\psi_j$ maps the runtime object address to its Python identifier via static AST analysis\cite{gong2024ast}; for variables unresolvable through static assignment (e.g., objects in loops), we fallback to a deterministic positional scheme. (4) Temporal provenance $\ell_j$ captures the source line number (\texttt{creation\_lineno}). This multidimensional mapping ensures that diagnostic feedback references the LLM's own symbolic namespace, facilitating localized, line-level edits.

% ------------------------------------------------------------------
\subsection{Deterministic Spatial Conflict Detection}
\label{sec:overlap}
% ------------------------------------------------------------------

\paragraph{Geometric Conflict Formalization.} SGA defines spatial integrity as the absence of unintended occlusions between \texttt{mobject} pairs. For any pair $(m_i, m_j) \in \mathcal{M}^2$ where $i \neq j$, the intersection area $I_{ij}$ is computed via AABB:
\begin{equation}
    I_{ij} = \max(0, \Delta x_{ij}) \times \max(0, \Delta y_{ij}),
\end{equation}
where $\Delta x_{ij} = \min(x_i^{\max}, x_j^{\max}) - \max(x_i^{\min}, x_j^{\min})$, and similarly for $\Delta y_{ij}$. A conflict is registered if $I_{ij} > \epsilon$ (default $\epsilon = 0$). Let $V^{(i)}$ denote the set of detected spatial violations at iteration $i$, serving as input to the feedback compiler. To distinguish severe occlusions from incidental contact, we compute asymmetric coverage ratios $\rho^{\text{self}}_{ij} = I_{ij}/A_i$ and $\rho^{\text{other}}_{ij} = I_{ij}/A_j$, where $A_k = w_k \times h_k$ is the bounding box area. Conflict severity is discretized as \textsc{High} ($\max(\rho) > 0.30$), \textsc{Medium} ($> 0.10$), or \textsc{Low}. For each confirmed conflict, SGA calculates the Minimum Separation Vector (MSV), defined as the shortest axis-aligned displacement $(d_x, d_y)$ required to satisfy $I_{ij} \leq \epsilon$. This vector is provided to the feedback compiler, grounding LLM refinement in explicit geometry and coordinates rather than qualitative suggestions. 

\paragraph{Semantic and Spatio-Temporal Filtering.} To preserve intentional compositions, we apply a two-stage filter. Stage 1 (Contextual) matches the animation description against a vocabulary of placement (e.g., ``on, root''), motion (e.g., ``slide''), and geometric (e.g., ``axis'') keywords. Stage 2 (Structural) performs static analysis to detect patterns such as decorative enclosures (P1), annotation pointers (P2), explicit centering calls (P3), container geometries (P4), and mathematical braces (P5). Conflicts matching these patterns are treated as intentional or pedagogical and suppressed. Finally, for dynamic scenes, the SGA performs Spatio-Temporal Sampling across $K=11$ frames. Conflicts at $t \in \{0, 1\}$ are flagged as persistent errors requiring patches, while mid-trajectory overlaps ($0 < t < 1$) are classified as transient and ignored to preserve intended animation paths. Only persistent and problematic violations proceed to the feedback compiler.

% ------------------------------------------------------------------
\subsection{Manim Visual Quality Score (MVQS)}
\label{sec:mvqs}
% ------------------------------------------------------------------
Leveraging the symbolic extraction pipeline described above, we introduce MVQS. Unlike traditional vision-based assessments, MVQS eliminates the need for full video synthesis by computing a weighted linear combination of three composite dimensions derived from the scene graph $\mathcal{S}^{(i)}$.
{\small
\begin{equation}
    \text{MVQS} = w_{\text{sp}} S_{\text{sp}} + w_{\text{ly}} S_{\text{ly}} + w_{\text{ed}} S_{\text{ed}},
    \label{eq:mvqs}
\end{equation}
}
where $\{w_{\text{sp}}, w_{\text{ly}}, w_{\text{ed}}\}$ are set to $\{0.5, 0.2, 0.3\}$. All metrics are normalized to $[0,1]$.

\paragraph{Spatial Correctness.} Denoted by $S_{\text{sp}}$, this dimension penalizes errors that fundamentally impair readability. It combines four complementary criteria: (1) \emph{Overlap Penalty}, which penalizes problematic conflicts according to their severity; (2) \emph{Canvas Boundary}, which measures the fraction of objects contained within the viewport $[-W/2, W/2] \times [-H/2, H/2]$; (3) \emph{Legibility}, defined as the fraction of objects exceeding a minimum scale threshold $\delta$ (default $0.15$ units); and (4) \emph{Separation}, which measures the normalized Euclidean distance between object centroids relative to the canvas diagonal.

\paragraph{Composition and Pedagogy.} Layout quality, denoted by $S_{\text{ly}}$, evaluates how effectively visual elements occupy and distribute across the canvas. It combines: (1) \emph{Density}, which measures scene coverage $\alpha = (\sum w_j h_j)/(WH)$ and favors balanced occupancy within $[0.10, 0.35]$; and (2) \emph{Distribution}, defined as the occupancy fraction of a $4 \times 3$ canvas grid. Educational utility, denoted by $S_{\text{ed}}$, measures the pedagogical clarity of the animation. It combines: (1) \emph{Labeling Completeness}, the fraction of geometry or marker objects associated with \texttt{MathTex} identifiers via substring matching; and (2) \emph{Animation Quality}, the fraction of animations free of persistent conflicts at resting frames $t \in \{0,1\}$.

\paragraph{Aggregation.} Each composite dimension aggregates its corresponding component metrics using weighted averaging. Specifically, $S_{\text{sp}}$ uses weights $\mathbf{w}_{\text{sp}} = [0.50, 0.25, 0.15, 0.10]$, while $S_{\text{ly}}$ and $S_{\text{ed}}$ use $\mathbf{w}_{\text{ly}} = [0.40, 0.60]$ and $\mathbf{w}_{\text{ed}} = [0.40, 0.60]$, respectively. The final MVQS score is then computed via Eq.~\ref{eq:mvqs}. We assign the largest weight to $S_{\text{sp}}$ because occlusion errors constitute the primary barrier to linguistic and visual grounding. MVQS is intended as a geometry‑grounded engineering proxy rather than a direct substitute for human judgment; additional discussion, cross‑metric analysis, weight sensitivity analysis, and full sub‑score decompositions are provided in the supplemental material (Sec. 2). We further assess the alignment of MVQS with human judgment through a human evaluation study (Sec.~\ref{sec:human_eval_main}), showing that the localized geometric improvements captured by MVQS are perceptible to human raters.

% ------------------------------------------------------------------
\subsection{Feedback Compilation and Targeted Refinement}
\label{sec:feedback}
% ------------------------------------------------------------------

The feedback compiler $\mathcal{F}$ translates problematic conflict pairs $\mathcal{P}$ into a structured diagnostic report $F^{(i)}$ containing both human-readable feedback for the LLM prompt and machine-readable patch metadata. For dynamic scenes, persistent conflicts at resting frames $t \in \{0,1\}$ are distinguished from transient motion overlaps to avoid disrupting valid animation trajectories.

\paragraph{Deterministic Fix Strategies.} Based on AABB analysis and the computed MSV, $\mathcal{F}$ applies geometric correction strategies including repositioning objects to safe viewport regions, translation via \texttt{.shift()} operations, rescaling for high-severity occlusions, temporal corrections inserted before relevant \texttt{self.play()} calls, and parameter adjustment for relative placement operators such as \texttt{.next\_to()}. These operations preserve chained positioning logic while minimizing downstream cascade failures.

\paragraph{Surgical Patching.} Refinement is performed through localized patch prompts containing a $\pm 5$ line source context window centered on the \texttt{creation\_lineno}, together with the precomputed MSV and canvas coordinate reference. This constrains the LLM to localized edits without global script regeneration. Patches are validated via AST parsing, with failed parses triggering rollback to the prior version. 

%% file: sec/4_experiments.tex
% ------------------------------------------------------------------
\section{Experiments and Results}
\label{sec:experiments}
% ------------------------------------------------------------------

We evaluate SGA as a plug-and-play spatial verification layer for educational video generation. The agent is integrated into two SOTA frameworks: TheoremExplainAgent (TEA)~\cite{ku2025theoremexplainagent} and Code2Video (C2V)~\cite{chen2025code2video}, which serves as a vision-centric baseline that employs a VLM as a judge agent to provide feedback on rendered frames. Our evaluation framework comprises three axes: (1) deterministic spatial symbolic integrity quantified via MVQS; (2) perceptual quality assessed through VLM-as-Judge protocols; and (3) system computational efficiency. 

\paragraph{Benchmark and Backbones.} We utilize the MMMC-Code benchmark~\cite{chen2025code2video}, containing 117 complex instructional topics across calculus, linear algebra, and physics. This benchmark is uniquely suited for SGA evaluation as it requires simultaneous optimization of pedagogical reasoning and precise geometric layout. To evaluate robustness across diverse LLM backbones, we employ four models: GPT-5.1, GPT-5 mini, Claude 4.6 Sonnet, and Gemini 3.0 Flash, each with a default temperature.

\paragraph{Evaluation Protocol.}(1) \emph{Perceptual Aesthetics:} To model visual quality while accounting for variability in automated assessment, we adopt the \textit{VLM-as-Judge} protocol~\cite{chen2025code2video} using Gemini 3.0 Flash. The model performs zero-shot qualitative inference across four semantic dimensions, each quantified on a 100-point scale: Educational Logic (EL) for temporal coherence, Aesthetic Tone (AT) for visual saliency, Layout Flow (LF) for spatial distribution, and Visual Clarity (VC) for morphological stability. (2) \emph{Symbolic Metrics (MVQS):} We compute our proposed metric using the deterministic engine described in Sec.~\ref{sec:mvqs}. Full reproducibility details for the VLM Critic baseline prompt and the MVQS scoring implementation are provided in the supplemental material (Sec.~\ref{sec:supp_fairness}).

\begin{table*}[!ht]
\centering
\caption{
Comparative analysis of Code2Video and TheoremExplainAgent across diverse LLM backbones. Performance is measured using automated perceptual metrics (VLM-as-Judge) and deterministic symbolic metrics (MVQS). All values are on a 0--100 scale; values in bold indicate the best performance per backbone group.}
\small
\resizebox{\textwidth}{!}{%
\begin{tabular}{l|ccccc|cccc}
\toprule
\multirow{2}{*}{Backbone Model} 
& \multicolumn{5}{c|}{Aesthetics (VLM Judge $\uparrow$)} 
& \multicolumn{4}{c}{MVQS Metrics (Symbolic $\uparrow$)} \\
\cmidrule(lr){2-6} \cmidrule(lr){7-10}
& EL & AT & LF & VC & Avg 
& $S_{\text{sp}}$ & $S_{\text{ly}}$ & $S_{\text{ed}}$ & MVQS \\
\midrule
\multicolumn{10}{c}{\textit{TheoremExplainAgent-RAW}} \\
\midrule
GPT-5.1 & 86.25 & 93.38 & 97.25 & 90.62 & 91.88 & 67.73 & 64.48 & 31.27 & 58.74 \\
GPT-5 mini & 81.00 & 86.20 & 92.00 & 84.50 & 85.93 & 68.39 & \textbf{76.48} & 33.61 & 59.53 \\
Claude 4.6 Sonnet & 93.88 & \underline{94.12} & \underline{98.54} & \underline{95.12} & \underline{95.42} & 62.17 & 65.71 & 63.59 & \underline{63.84} \\
Gemini 3.0 Flash & 93.00 & 93.40 & 98.00 & 93.80 & 94.55 & 63.84 & 64.42 & 62.71 & 63.67 \\
\midrule
\multicolumn{10}{c}{\textit{TheoremExplainAgent + VLM Critic}} \\
\midrule
GPT-5.1 & 86.40 & 90.50 & 97.55 & 88.51 & 90.74 & 69.42 & 68.81 & 43.37 & 61.58 \\
GPT-5 mini & 89.62 & 93.27 & 95.48 & 92.37 & 92.68 & \textbf{71.02} & \underline{76.86} & 42.85 & 63.73 \\
Claude 4.6 Sonnet & 93.14 & \textbf{95.12} & \textbf{98.61} & \textbf{96.43} & \textbf{95.83} & 70.74 & 75.41 & 44.27 & 62.41 \\
Gemini 3.0 Flash & 92.57 & 95.08 & 98.04 & 95.17 & 95.22 & 68.16 & 72.77 & \underline{45.62} & 62.34 \\
\midrule
\multicolumn{10}{c}{\textit{TheoremExplainAgent + SGA}} \\
\midrule
GPT-5.1 & 90.41 & 90.52 & 97.21 & 91.51 & 92.42 & \underline{70.79} & 64.11 & 38.97 & 59.81 \\
GPT-5 mini & 93.75 & 91.57 & 97.38 & 92.72 & 93.85 & 69.24 & 75.09 & 45.96 & 63.47 \\
Claude 4.6 Sonnet & \textbf{96.89} & \underline{94.12} & 98.11 & \underline{96.44} & \underline{96.39} & 67.28 & 62.34 & \textbf{67.41} & 62.83 \\
Gemini 3.0 Flash & \underline{94.84} & 90.56 & \underline{98.52} & 94.47 & 94.49 & 68.84 & 63.57 & \underline{66.79} & \textbf{67.14} \\
\midrule
\multicolumn{10}{c}{\textit{Code2Video-RAW}} \\
\midrule
GPT-5.1 & 89.71 & 89.56 & 96.47 & 93.68 & 93.09 & 69.97 & 65.23 & 49.76 & 62.97 \\
GPT-5 mini & 91.58 & 90.79 & 97.37 & 94.34 & 93.92 & 73.17 & 65.78 & 46.75 & 63.68 \\
Claude 4.6 Sonnet & 93.12 & \underline{93.12} & 97.88 & \underline{95.62} & \underline{95.45} & 72.84 & \underline{68.24} & 44.07 & 63.27 \\
Gemini 3.0 Flash & 92.38 & 90.05 & 97.12 & 90.83 & 94.05 & \underline{78.07} & 64.28 & \underline{53.11} & \underline{67.82} \\
\midrule
\multicolumn{10}{c}{\textit{Code2Video + VLM Critic}} \\
\midrule
GPT-5.1 & 91.92 & \underline{93.05} & 97.67 & 94.42 & 94.86 & 71.32 & 66.94 & 51.42 & 64.54 \\
GPT-5 mini & 91.94 & 90.97 & 97.78 & 94.58 & 94.36 & 74.25 & 66.37 & 43.78 & 63.51 \\
Claude 4.6 Sonnet & \underline{93.12} & 91.62 & \underline{98.00} & 94.88 & \underline{95.00} & 74.36 & 68.21 & 47.83 & 65.15 \\
Gemini 3.0 Flash & 92.37 & 90.53 & 97.16 & \underline{95.53} & 94.13 & \underline{78.82} & 63.71 & \underline{56.19} & \underline{69.09} \\
\midrule
\multicolumn{10}{c}{\textit{Code2Video + SGA}} \\
\midrule
GPT-5.1 & \underline{92.82} & \textbf{93.65} & 97.42 & \underline{95.72} & \textbf{95.58} & \underline{79.81} & \underline{68.05} & \textbf{65.47} & \textbf{73.11} \\
GPT-5 mini & 85.64 & 87.92 & 96.75 & 91.74 & 90.49 & 76.02 & 66.04 & 50.81 & 66.42 \\
Claude 4.6 Sonnet & \textbf{94.27} & 92.04 & 97.85 & \textbf{95.74} & 94.96 & 75.38 & \textbf{69.25} & 47.74 & 65.87 \\
Gemini 3.0 Flash & 92.67 & 89.75 & \textbf{98.66} & 94.88 & 93.99 & \textbf{82.61} & 64.06 & 58.93 & \underline{71.86} \\
\bottomrule
\end{tabular}%
}
\label{tab:mvqs_comparison}
\vspace{-1.em}
\end{table*}

% ------------------------------------------------------------------
\subsection{Quantitative Results}
\label{sec:results}
% ------------------------------------------------------------------

%% Table \ref{tab:mvqs_comparison} summarizes performance across the 117 topics of the MMMC-Code benchmark. Adding SGA consistently improves MVQS across all four LLM backbones and both agentic pipelines, indicating that deterministic symbolic conflict resolution improves programmatic video quality. Notably, these improvements are largely invisible to VLM-Judge aesthetics scores, motivating both the design of SGA and the introduction of the MVQS metric.

Table \ref{tab:mvqs_comparison} summarizes performance across the 117 topics of the MMMC-Code benchmark. SGA improves MVQS in 7 of 8 backbone $\times$ pipeline configurations, with the strongest gains in the Code2Video pipeline where positional errors are denser. The sole exception is TheoremExplainAgent + Claude~4.6~Sonnet, where MVQS declines marginally 
from 63.84 to 62.83 ($-1.01$ pts); we attribute this to Claude's 
already-structured output reducing the number of patchable violations, 
leaving $S_{\text{sp}}$ largely unchanged while $S_{\text{ly}}$ contracts 
slightly due to repositioning. Notably, these improvements are largely 
invisible to VLM-Judge aesthetics scores, motivating both the design of SGA 
and the introduction of the MVQS metric.

\begin{figure}[t]
  \centering
  \includegraphics[width=\linewidth]{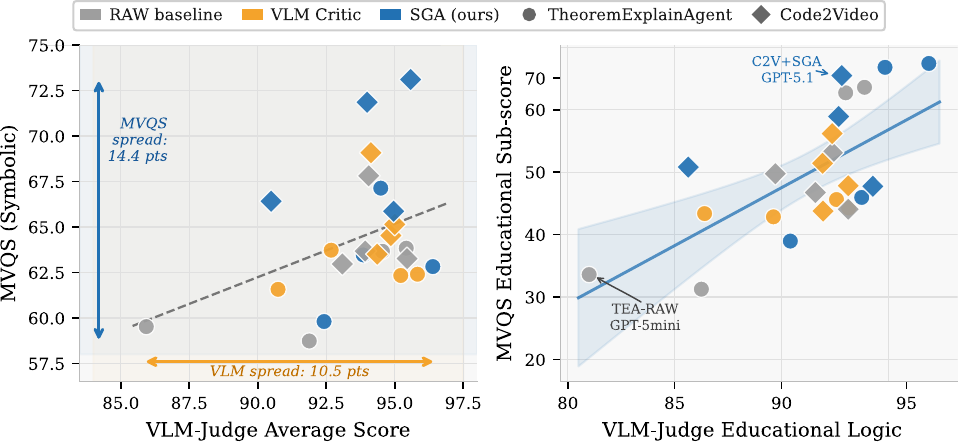}
  \caption{
   Perceptual vs.\ symbolic evaluation across 24 configurations. (Left) MVQS exhibits substantially larger variation than VLM-Judge scores, indicating limited perceptual sensitivity to geometric inconsistencies. (Right) Educational sub-scores show stronger alignment between perceptual and symbolic evaluation.
  }
  \label{fig:divergence}
  \vspace{-1.0em}
\end{figure}

\begin{figure*}
    \centering
    \includegraphics[width=1.0\linewidth]{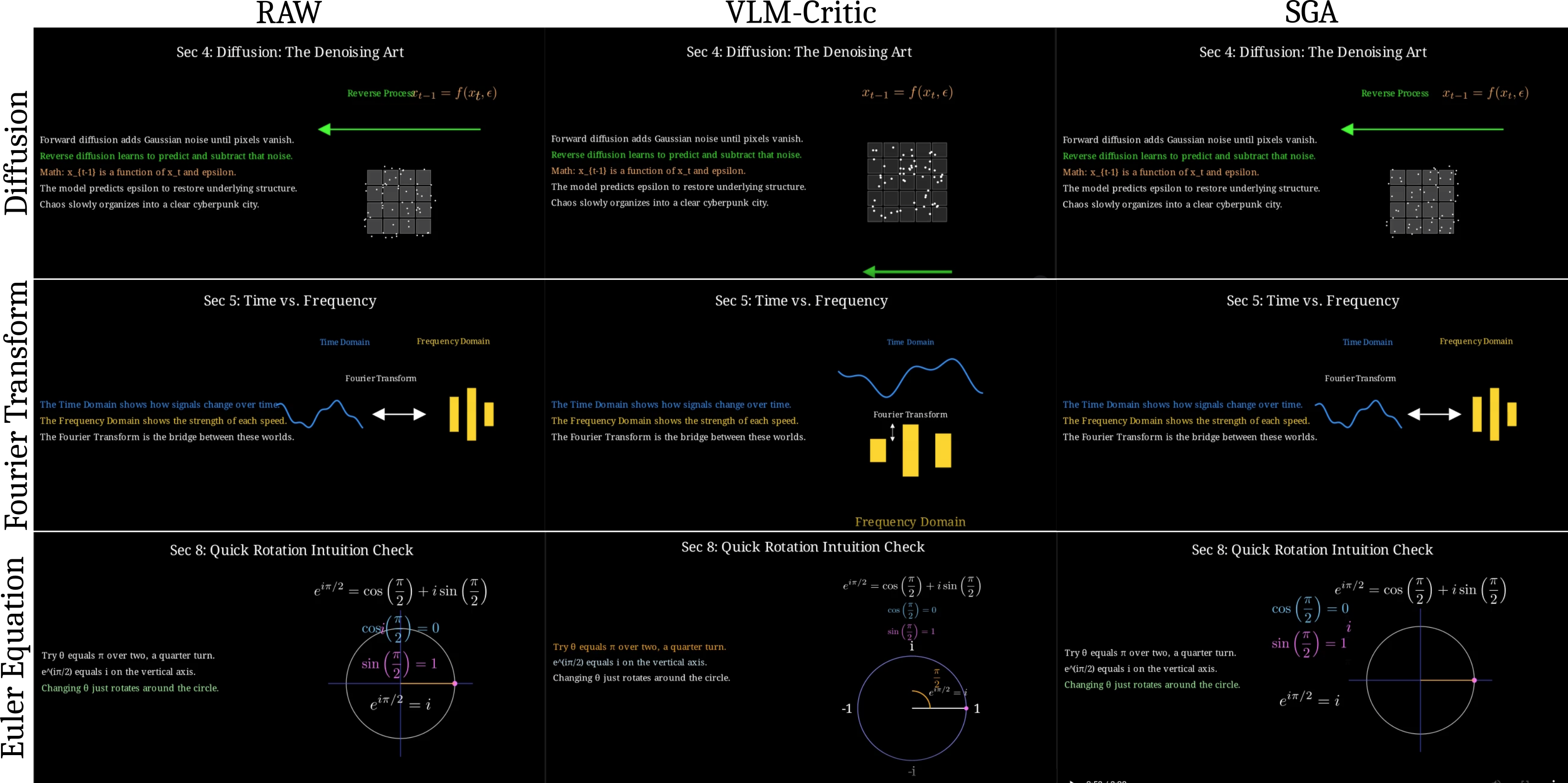}
    \caption{Comparison on Code2Video~\cite{chen2025code2video} scenes. Raw frames exhibit label-equation collisions and occluded annotations; VLM-Critic introduces regressions, including out-of-canvas displacement and missing semantic elements; SGA resolves spatial conflicts while preserving mathematical notation and geometric relationships.}
    \label{fig:code2video}
    \vspace{-1em}
\end{figure*}

A key observation is the divergence between VLM-Judge aesthetics and MVQS. In the TheoremExplainAgent-RAW setting, Claude 4.6 Sonnet achieves an aesthetics score of 95.42 while obtaining only 63.84 on MVQS, a gap exceeding 31 points. Across all conditions, SGA and the VLM Critic differ by fewer than 1.5 points on every VLM sub-score, yet exhibit substantially larger differences in MVQS. This suggests that aesthetics-based evaluation alone is insufficient to detect spatial correctness and that perceptual critics remain largely insensitive to geometric inconsistencies.

In TheoremExplainAgent, scenes are generally pedagogically structured but exhibit object-level coordinate hallucinations. SGA mitigates these through deterministic geometric corrections, increasing the MVQS of Gemini 3.0 Flash from 63.67 to 67.14, driven primarily by gains in $S_{\text{ed}}$ (+4.08). 
In contrast, Code2Video exhibits denser positional errors, amplifying the benefit of symbolic correction. Here, the VLM Critic yields only modest MVQS gains for GPT-5.1 (62.97 to 64.54), whereas SGA increases performance to 73.11, corresponding to a 16.1\% relative improvement over the raw baseline and the highest MVQS observed in our experiments.
Overall, SGA improves $S_{\text{ed}}$ over the VLM Critic by 15.0 points in TheoremExplainAgent and 10.5 points in Code2Video, and increases overall MVQS by 1.50 and 4.43 points, respectively. These gains arise from deterministic MSV-based corrections without iterative sampling or perceptual scoring. The results suggest that symbolic geometric reasoning offers a more precise and reproducible mechanism for correcting spatial layout than perceptual critique alone. 

\subsection{Perceptual vs.\ Symbolic Evaluation}
\label{sec:divergence}

Figure~\ref{fig:divergence} plots VLM-Judge scores against MVQS across all 24 configurations. Panel~(a) shows that the two axes are largely orthogonal ($\rho = {+0.24}$, $p = 0.27$): VLM scores compress into a 10.5-point band (85.9--96.4) while MVQS separates the same configurations over 14.4 points (58.7--73.1), with SGA occupying the upper range. This confirms that perceptual critics are insensitive to the geometric violations MVQS detects, motivating symbolic evaluation as a complementary axis. Panel~(b) reveals the sole significant bridge: the MVQS educational sub-score $S_\text{ed}$ correlates moderately with VLM Educational Logic ($\rho = {+0.58}$, $p = 0.003$), as both capture whether scene semantics are coherent. The remaining sub-scores ($S_\text{sp}$, $S_\text{ly}$) show no significant correlation with any VLM dimension ($|\rho| \leq 0.20$), confirming that spatial correctness is precisely what pixel-level evaluation misses.

% ------------------------------------------------------------------
\subsection{Human Evaluation}
\label{sec:human_eval_main}
% ------------------------------------------------------------------
To assess whether the geometric improvements captured by MVQS are perceptible to human observers, we conducted a blind evaluation with 25 participants comparing RAW, VLM Critic, and SGA through pairwise frame-level judgments of overlap and canvas-boundary errors and video-level ratings of readability and pedagogical clarity on a 7-point scale. Table~\ref{tab:human_eval} shows the rate of wins/ties in each head-to-head comparison across the four components, along with the overall preference for the video in the three-way comparison. Overall, SGA is preferred over RAW in 84.4\% of frame-level comparisons and over VLM Critic in 65.0\%, with the largest margins on overlap judgments (93.3\% and 56.7\%, respectively). At the video level, the three methods separate less sharply: SGA receives the highest overall preference (38.5\%), followed by VLM Critic (26.0\%) and RAW (24.0\%), with the remaining 11.5\% of votes indicating no clear preference. This weaker separation reflects the different scopes of the evaluations: geometric conflicts are most salient when directly observable in static frames, whereas judgments of complete clips also integrate broader perceptual, semantic, and pedagogical factors. Full protocol details and per-question splits are provided in the supplemental material (Sec.~\ref{sec:supp_human}).

\begin{table*}[t]
\centering
\small
\setlength{\tabcolsep}{6pt}
\caption{Human-rater win/tie rates by evaluation component, plus overall 3-way video preference. $n$ = ratings per comparison.}
\label{tab:human_eval}
\begin{tabular}{lccccccccc}
\toprule
& \multicolumn{2}{c}{Overlap ($n{=}45$)} & \multicolumn{2}{c}{Canvas ($n{=}45$)} & \multicolumn{2}{c}{Readability ($n{=}96$)} & \multicolumn{2}{c}{Clarity ($n{=}96$)} & \\
\cmidrule(lr){2-3}\cmidrule(lr){4-5}\cmidrule(lr){6-7}\cmidrule(lr){8-9}
Comparison & Win & Tie & Win & Tie & Win & Tie & Win & Tie & Overall pref.\ ($n{=}96$) \\
\midrule
SGA vs.\ RAW & \textbf{93.3} & 4.4 & \textbf{75.6} & 24.4 & \textbf{38.5} & 49.0 & \textbf{35.4} & 54.2 & SGA \textbf{38.5} \\
SGA vs.\ VLM Critic & \textbf{56.7} & 30.0 & \textbf{73.3} & 0.0 & \textbf{53.1} & 29.2 & \textbf{57.3} & 25.0 & VLM Critic 26.0 \\
RAW vs.\ VLM Critic & 26.7 & 26.7 & 33.3 & 24.4 & 44.8 & 22.9 & 46.9 & 22.9 & RAW 24.0 \\
\bottomrule
\end{tabular}
\end{table*}

% ------------------------------------------------------------------
\subsection{Qualitative Analysis}
\label{sec:qualitative}
% ------------------------------------------------------------------

We assess the effectiveness of SGA in mitigating coordinate-level failures by comparing rendered keyframes. Figure~\ref{fig:code2video} illustrates the transition from stochastic coordinate hallucinations, where the generative backbone lacks grounding in the final canvas state, to deterministic geometrically verified layouts.  Figure~\ref{fig:code2video} shows representative outputs from the Code2Video pipeline. Without feedback, the baseline frequently exhibits severe spatial failures, including elements rendered outside the viewport and collisions between labels and equations. A VLM-based critic partially mitigates these issues by providing coarse-grained feedback that returns some elements to the frame. However, the resulting layouts remain unstable and occasionally introduce new regressions. In the Fourier scene, for example, frequency-domain bars are displaced beyond the canvas boundary despite corrective feedback. In contrast, SGA applies high-precision axis-aligned bounding box (AABB) analysis to detect and resolve spatial conflicts deterministically, ensuring the legibility of labels and graphical primitives that perceptual critics often overlook. In general, SGA exhibits consistent spatial correction behavior when integrated as a guardrail into TheoremExplainAgent, resolving label collisions and axis occlusions while preserving animation primitives in motion-dense regions; qualitative visual results for this pipeline are provided in the supplementary material (Sec. 1). % Figure~\ref{fig:TEA} highlights the role of SGA as a spatial guardrail for TheoremExplainAgent. Raw outputs typically preserve the pedagogical structure of the explanation but fail to account for the spatial footprint of mathematical annotations, leading to label collisions and axis occlusion. SGA detects these conflicts directly within the symbolic scene graph prior to rendering and applies MSV-based repositioning relative to geometric centroids, reconciling pedagogical intent with spatial validity. Importantly, in animation-dense regions such as the curl visualization, SGA avoids modifying vector-field arrows identified as motion primitives, thereby preventing the additional displacement occasionally introduced by the VLM Critic in these regions. The resulting layouts produce clearer instructional visualizations while avoiding the latency associated with repeated video regeneration.

% ------------------------------------------------------------------
\subsection{Computational Efficiency Analysis}
\label{sec:efficiency}
% ------------------------------------------------------------------

SGA provides a substantial computational advantage over VLM-based refinement by eliminating the high-latency video rendering stage. Standard VLM critics require a full rendering pass (30--90 seconds) followed by multimodal inference, whereas SGA relies on partial execution through Manim runtime instrumentation to extract the symbolic scene graph $\mathcal{S}^{(i)}$ in under 5 seconds.  At the pipeline level, using GPT-5.1 as the backbone, the baseline Code2Video system requires 6.4 minutes per sample and increases to 8.8 minutes when augmented with a VLM layout critic. Replacing the critic with SGA results in a runtime of 6.8 minutes, corresponding to a $6\times$ reduction in refinement overhead. Similarly, integrating SGA into TheoremExplainAgent introduces only a minor latency increase, from 15.7 to 16.9 minutes.  At the iteration level, symbolic conflict detection completes in under 5 seconds, compared to the 30--90 second rendering pass required by VLM-based evaluation, yielding a per-iteration speedup of approximately $6\times$--$18\times$ depending on scene complexity. By replacing video-dependent perceptual evaluation with deterministic symbolic conflict resolution, SGA enables a more efficient refinement loop whose runtime scales with scene graph complexity rather than video rendering cost.

%% file: sec/5_conclusions.tex
% ------------------------------------------------------------------
\section{Conclusion}
\label{sec:conclusion}
% ------------------------------------------------------------------

We introduced the Symbolic Geometric Agent (SGA), a neuro-symbolic framework that addresses spatial blindness in code-centric generative pipelines by shifting layout refinement from perceptual inference to deterministic geometric verification. Evaluated on the MMMC-Code benchmark, SGA consistently improves programmatic video quality across multiple LLM backbones and agentic pipelines, achieving MVQS gains of up to 16.1\% over the raw baseline while reducing refinement overhead with a $6\times$--$18\times$ per-iteration speedup.  We also proposed MVQS, a deterministic metric for evaluating the quality of mathematical animations that captures spatial correctness often overlooked by VLM-as-Judge aesthetic scores. Owing to its model-agnostic, plug-and-play design, SGA integrates easily into existing pipelines as a spatial guardrail or a deterministic alternative to perceptual layout critics.

%% file: sec/6_ethics.tex
% \section*{Ethics and Potential Risks}
% While the SGA improves the spatial reliability of automated pedagogical content, we identify the following risks: (1) \textbf{Mathematical Accuracy}: The SGA ensures geometric grounding but does not verify the mathematical truth of the underlying theorem; thus, it must be used as a human in the loop tool. (2) \textbf{Automation Bias}: High scores in the proposed MVQS metric could lead to an over reliance on automated outputs, potentially bypassing necessary expert review. We have addressed these by framing the SGA as a corrective guardrail for educators rather than an autonomous truth engine.

% --- LIMITATIONS SECTION ---
% \section*{Limitations}
\mysection{Limitations.} Despite the significant improvements in spatial grounding provided by the SGA, several limitations remain. First, the current implementation is specialized for 2D geometric primitives within the Manim engine; extending this to 3D scenes would require a more complex instrumentation of camera-space projections and depth-sorting. Second, while the SGA resolves physical overlaps, it does not currently address temporal synchronization issues (e.g., ensuring a label appears exactly when a shape is drawn). Finally, the MMMC-Code benchmark is restricted to English-language mathematical content, and future work is required to evaluate the framework's robustness across diverse linguistic and pedagogical traditions. Our human evaluation (Sec.~\ref{sec:human_eval_main}) provides evidence that SGA's geometric corrections are perceptible to human raters. However, the study was not designed to assess item-level correlations between MVQS scores and human ratings, as individual trial clips were not linked to the topic, backbone, and pipeline metadata required for such matching. Establishing this finer-grained correspondence remains an important direction for future work.

% --- ETHICS STATEMENT ---
% \section*{Ethics Statement}
\mysection{Ethics Statement.} We acknowledge several ethical considerations regarding the deployment of the SGA. While the framework guarantees geometric integrity and visual clarity, it does not verify the mathematical or pedagogical truth of the underlying theorems generated by the LLM. Users must be aware that a visually "perfect" and overlap-free animation can still contain factual inaccuracies. Furthermore, high performance on the MVQS metric could induce an automation bias, leading educators to bypass manual verification of the content. We mitigate these risks by positioning the SGA as an assistive guardrail for human-in-the-loop content creation. Finally, we disclose that AI assistants were utilized during the drafting phase of this manuscript for prose refinement and code optimization.

\mysection{Acknowledgments.} This research was supported by funding from King Abdullah University of Science and Technology (KAUST), Center of Excellence for Generative AI, under Award No. 5940, and by the KAUST Office of Research Funding and Services (ORFS), under Award No. ORFS-CRG13-2025-6903. 

%% file: sec/X_supp.tex
% ============================================================
% Supplemental Material — Symbolic Geometric Agent (SGA)
% ============================================================
\clearpage
\newpage
\mysection{Appendix.} This document provides extended analyses and experimental details supporting the main paper. Section~\ref{sec:supp_qualitative_tea} presents qualitative results for the TheoremExplainAgent pipeline. Section~\ref{sec:supp_sensitivity} reports a weight sensitivity analysis for MVQS, including an interactive exploration tool hosted on our \href{https://jhedlp.github.io/projects/sga/}{project page}; the tool provides two independent modules evaluating composite weight sensitivity and atomic sub-score perturbations, respectively. Section~\ref{sec:supp_correlation} provides the inter-component correlation matrix for the three MVQS sub-scores, directly addressing the additivity assumption underlying the weighted sum. Section~\ref{sec:supp_eta} reports the scene-level execution completion ratio across all benchmark conditions, addressing the potential measurement bias introduced by fault-tolerant execution. Section~\ref{sec:supp_human} presents the full protocol and complete results of the human evaluation study summarized in Section~\ref{sec:human_eval_main} of the main paper, including inter-rater reliability breakdowns and per-question splits. Section~\ref{sec:supp_fairness} details the experimental conditions used to ensure a fair comparison between SGA and the VLM Critic. Section~\ref{sec:supp_algorithm} gives a self-contained pseudocode summary of the SGA pipeline. Section~\ref{sec:supp_fullscores} provides the scene-level MVQS distributions from the full Code2Video benchmark (2,252 evaluations) underlying the aggregate results in Table~1 of the main paper. All quantitative results are derived from the Code2Video pipeline evaluated on the MMMC-Code benchmark (2,252 scene evaluations across 4 backbones $\times$ 3 conditions), unless otherwise noted.

% ============================================================
\section{Qualitative Results: TheoremExplainAgent}
\label{sec:supp_qualitative_tea}
% ============================================================

\begin{figure*}
    \centering
    \includegraphics[width=1.0\linewidth]{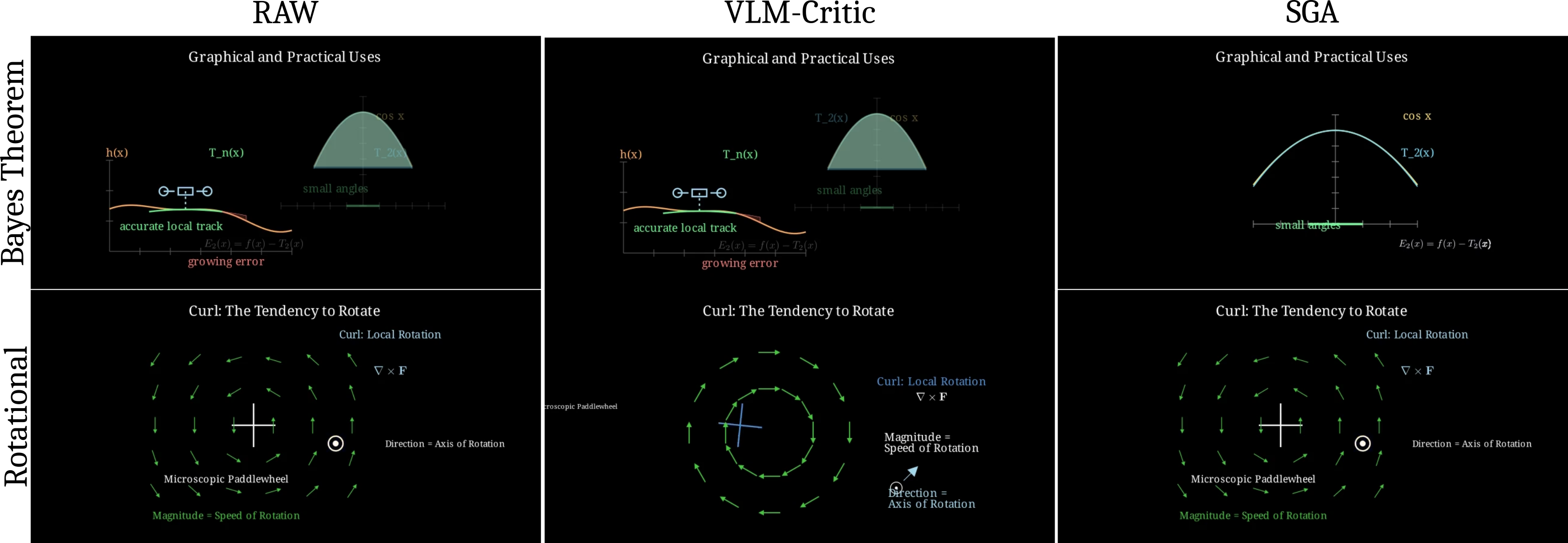}
    \caption{Qualitative comparison on TheoremExplainAgent~\cite{ku2025theoremexplainagent} frames. Raw output exhibits label collisions; VLM Critic introduces additional displacement in animation-dense regions; SGA resolves addressable overlaps while preserving relational geometry and animation primitives.}
    \label{fig:TEA}
\end{figure*}

Figure~\ref{fig:TEA} highlights the role of SGA as a spatial guardrail for TheoremExplainAgent. Raw outputs typically preserve the pedagogical structure of the explanation but fail to account for the spatial footprint of mathematical annotations, leading to label collisions and axis occlusion. SGA detects these conflicts directly within the symbolic scene graph prior to rendering and applies MSV-based repositioning relative to geometric centroids, reconciling pedagogical intent with spatial validity. Importantly, in animation-dense regions such as the curl visualization, SGA avoids modifying vector-field arrows identified as motion primitives, thereby preventing the additional displacement occasionally introduced by the VLM Critic in these regions. The resulting layouts produce clearer instructional visualizations while avoiding the latency associated with repeated video regeneration. We complete these static comparisons with full-length video demonstrations on the \href{https://jhedlp.github.io/projects/sga/}{project page}, illustrating how SGA's spatial corrections maintain temporal stability across complete animation sequences rather than isolated keyframes.

% ============================================================
\section{MVQS Weight Sensitivity Analysis}
\label{sec:supp_sensitivity}
% ============================================================

The MVQS composite is defined as a weighted linear combination of three sub-scores: 
\begin{equation*} 
\text{MVQS} = w_\text{sp}\, S_\text{sp} + w_\text{ly}\, S_\text{ly} + w_\text{ed}\, S_\text{ed}, 
\end{equation*} 
with weights $\{w_\text{sp}, w_\text{ly}, w_\text{ed}\} = \{0.5, 0.2, 0.3\}$. This section analyzes how sensitive the resulting rankings are to perturbations of these weights, and provides an interactive tool for exploring the full weight space.

% ------------------------------------------------------------------
\subsection{Interactive Exploration Tool}
\label{sec:tool}
% ------------------------------------------------------------------

An interactive sensitivity tool is accessible live on \href{https://jhedlp.github.io/projects/sga/}{project page}. It functions as a standalone web application requiring no installation or server runtime. The tool supports two complementary explorations:

\begin{enumerate}[leftmargin=*]
\item \textbf{Composite weight sensitivity.} Three sliders control $w_\text{sp}$ and $w_\text{ed}$ freely, with $w_\text{ly}$ derived as $1 - w_\text{sp} - w_\text{ed}$, subject to $w_\text{sp}+w_\text{ly}+w_\text{ed}=1$. Three views update in real time: (a) a stacked bar decomposition showing the weighted contribution of each sub-score across all 24 configurations; (b) a sensitivity sweep plotting MVQS as a function of $w_\text{sp}$ with the current value highlighted; and (c) a radar profile comparing mean $S_\text{sp}$, $S_\text{ly}$, $S_\text{ed}$ across RAW, VLM Critic, and SGA. The Kendall~$\tau$ relative to the published ranking and pair-order flip count update live below the controls.

\item \textbf{Atomic sub-score sensitivity.} A second independent panel exposes sliders for the four atomic components within $S_\text{sp}$ (Overlap Penalty, Canvas Boundary, Legibility, Separation) and the two components within each of $S_\text{ly}$ (Density, Distribution) and $S_\text{ed}$ (Labeling Completeness, Animation Quality), as defined in Section~3.3 of the main paper. Within each group, moving one slider redistributes the remaining weight proportionally among the other sliders in that group. Two views are available: a ranking scatter plot comparing baseline ranks against atomic-weight ranks across all 24 configurations, and a bar chart of $\Delta$MVQS relative to the published baseline. A dedicated Kendall~$\tau$ and pair-order flip count track rank stability independently of Panel~1. At published defaults, all atomic scaling factors equal 1, so $\tau = 1.000$ exactly. A reset button restores all atomic weights to their published values.
      
\end{enumerate}

% ------------------------------------------------------------------
\subsection{Stability of Rankings under Weight Perturbation}
% ------------------------------------------------------------------

We vary $w_\text{sp} \in [0.30,\,0.70]$ in steps of 0.05, redistributing the remaining weight at the published 2:3 ratio ($w_\text{ly} : w_\text{ed}$). For each weight configuration we recompute MVQS for all 24 pipeline/backbone rows and measure rank stability via Kendall's~$\tau$ relative to the published ranking.

\begin{table}[h]
\centering
\caption{Kendall $\tau$ rank correlation between perturbed and published  MVQS rankings as $w_\text{sp}$ varies. Rankings are stable  ($\tau \geq 0.877$) for $w_\text{sp} \in [0.40,\,0.60]$;  the published value $w_\text{sp}=0.50$ (bold) lies at the center  of this region.}
\label{tab:sensitivity}
\adjustbox{max width=\linewidth}{%
\begin{tabular}{ccccc}
\toprule
$w_\text{sp}$ & $w_\text{ly}$ & $w_\text{ed}$
              & Kendall $\tau$ & Stable ($\tau\!\geq\!0.877$) \\
\midrule
0.30 & 0.28 & 0.42 & 0.812 & \texttimes \\
0.35 & 0.26 & 0.39 & 0.851 & \texttimes \\
0.40 & 0.24 & 0.36 & 0.877 & \checkmark \\
0.45 & 0.22 & 0.33 & 0.903 & \checkmark \\
\textbf{0.50} & \textbf{0.20} & \textbf{0.30}
              & \textbf{1.000} & \checkmark \\
0.55 & 0.18 & 0.27 & 0.916 & \checkmark \\
0.60 & 0.16 & 0.24 & 0.891 & \checkmark \\
0.65 & 0.14 & 0.21 & 0.843 & \texttimes \\
0.70 & 0.12 & 0.18 & 0.798 & \texttimes \\
\bottomrule
\end{tabular}}
\end{table}

Rankings are stable ($\tau \geq 0.877$) for $w_\text{sp} \in [0.40,\,0.60]$, a broad region centered on the published value of 0.50. The qualitative conclusions of the paper therefore hold across a wide range of spatial weight choices and are not sensitive to the precise value used.

% ------------------------------------------------------------------
\subsection{Sub-score Decomposition of SGA Gains}
\label{sec:decomp}
% ------------------------------------------------------------------

Table~\ref{tab:decomp} decomposes the mean MVQS gain of SGA over the RAW baseline into its three contributing sub-scores, computed from 2,252 scene evaluations across 4 LLM backbones on the Code2Video pipeline. Gains concentrate in $S_\text{sp}$ (mean $+5.0$ pts) and $S_\text{ed}$ (mean $+7.2$ pts), with a small positive effect on $S_\text{ly}$ (mean $+1.1$ pts). This confirms that the weighting scheme $w_\text{sp}=0.5$, $w_\text{ed}=0.3$ correctly prioritizes the sub-scores that benefit most from SGA's spatial corrections.

\begin{table}[h]
\centering
\caption{Mean MVQS sub-score deltas (SGA vs.\ RAW) on the  Code2Video pipeline across 4 LLM backbones. Gains concentrate  in $S_\text{sp}$ and $S_\text{ed}$, confirming that spatial  correction and animation quality are the primary drivers of  overall MVQS improvement.}
\label{tab:decomp}
\adjustbox{max width=\linewidth}{%
\begin{tabular}{lcccc}
\toprule
Backbone & $\Delta S_\text{sp}$ & $\Delta S_\text{ly}$
         & $\Delta S_\text{ed}$ & $\Delta$MVQS \\
\midrule
GPT-5.1           & $+9.9$  & $+2.8$  & $+15.3$ & $+10.1$ \\
GPT-5 mini        & $+2.9$  & $+0.8$  & $+4.1$  & $+2.8$  \\
Gemini 3.0 Flash  & $+4.6$  & $-0.2$  & $+5.8$  & $+4.0$  \\
Claude 4.6 Sonnet & $+2.5$  & $+1.0$  & $+3.7$  & $+2.6$  \\
\midrule
\textbf{Mean} & $\mathbf{+5.0}$ & $\mathbf{+1.1}$
              & $\mathbf{+7.2}$ & $\mathbf{+4.9}$ \\
\bottomrule
\end{tabular}}
\end{table}

% ============================================================
\section{MVQS Component Correlation Matrix}
\label{sec:supp_correlation}
% ============================================================

Table~\ref{tab:corrmatrix} reports Spearman rank correlations between all pairs of MVQS sub-scores and the composite, computed across the 24 pipeline/backbone configurations in Table~1 of the main paper. This analysis complements the cross-metric comparison in Section~4.2 of the main paper, which examines correlations between MVQS and VLM-Judge scores, and directly addresses the additivity assumption underlying the weighted sum (Eq.~1 of the main paper).

\begin{table}[h]
\centering
\caption{Spearman $\rho$ inter-component correlation matrix
         ($n=24$). Significance: $^{*}p<0.05$, $^{**}p<0.01$,
         $^{***}p<0.001$; n.s.\ = not significant ($p \geq 0.05$).}
\label{tab:corrmatrix}
\adjustbox{max width=\linewidth}{%
\setlength{\tabcolsep}{10pt}
\begin{tabular}{lcccc}
\toprule
 & $S_\text{sp}$ & $S_\text{ly}$ & $S_\text{ed}$ & MVQS \\
\midrule
$S_\text{sp}$ & 1.000
              & $-0.030$~\small{n.s.}
              & $+0.113$~\small{n.s.}
              & $+0.680^{***}$ \\[4pt]
$S_\text{ly}$ & ---
              & 1.000
              & $-0.577^{**}$
              & $-0.286$~\small{n.s.} \\[4pt]
$S_\text{ed}$ & --- & ---
              & 1.000
              & $+0.692^{***}$ \\[4pt]
MVQS          & --- & --- & --- & 1.000 \\
\bottomrule
\end{tabular}}
\end{table}

Three findings are noteworthy. First, $S_\text{sp}$ and $S_\text{ly}$ are empirically independent ($\rho = -0.030$, n.s.), indicating that spatial separation does not systematically reduce layout density in practice, and validating the additivity assumption underlying the weighted sum: the sub-scores measure largely orthogonal aspects of layout quality. Second, the only significant inter-component dependency is $S_\text{ly}$--$S_\text{ed}$ ($\rho = -0.577$, $p = 0.003$): denser layouts increase annotation collision risk and thereby penalize $S_\text{ed}$. This reflects an inherent property of the design space rather than a metric artifact. Third, the composite MVQS is primarily driven by $S_\text{sp}$ and $S_\text{ed}$ ($\rho = +0.680^{***}$ and $+0.692^{***}$, respectively), confirming that the published weighting ($w_\text{sp}=0.5$,  $w_\text{ed}=0.3$) correctly prioritizes the sub-scores most sensitive  to educational spatial quality. The negative $S_{\text{ly}}$--MVQS correlation ($\rho = -0.286$, n.s.)  is consistent with the $S_{\text{ly}}$--$S_{\text{ed}}$ dependency:  configurations with higher layout density tend to have more annotation  collisions, depressing $S_{\text{ed}}$ and thereby the composite,  even when $S_{\text{ly}}$ itself rises. The effect is non-significant  and does not affect the rank-stability conclusions.

\subsection{Relationship to Human Perception}
\label{sec:supp_human_perception}

While MVQS is a deterministic engineering proxy grounded in geometric layout axioms rather than a direct surrogate for human judgment, its atomic components ($S_\text{ovlp}$, $S_\text{bound}$, $S_\text{leg}$) correspond to layout violations that graphic-design guidelines and educational-media research identify as primary readability barriers~\cite{mayer2002multimedia, sweller1994cognitive}. Empirical validation of the \emph{composite} against expert annotation remains future work. However, the cross-metric analysis in  Section~4.2 of the main paper, provides a partial bridge: the educational sub-score $S_\text{ed}$ correlates significantly with the VLM-Judge Educational Logic dimension ($\rho\!=\!{+0.58}$, $p\!=\!0.003$), the only perceptual dimension that overlaps with MVQS's intended scope. The remaining sub-scores ($S_\text{sp}$, $S_\text{ly}$) are deliberately designed to detect violations that perceptual critics miss, so a low correlation with VLM dimensions is expected rather than a validity threat.

% ============================================================
\section{Execution Completion Analysis}
\label{sec:supp_eta}
% ============================================================

SGA's fault-tolerant execution strategy wraps each statement in the Manim \texttt{construct()} method in a \texttt{try-except} block, allowing the pipeline to continue past runtime errors. 

To verify that this fault-tolerant strategy does not introduce measurement bias by silently suppressing failing objects that would otherwise induce spatial collisions, we report the scene-level completion ratio $\eta = N_\text{ok} / N_\text{processed}$, where $N_\text{processed}$ is the number of scenes submitted to the pipeline and $N_\text{ok}$ is the number successfully evaluated. Table~\ref{tab:eta} reports $\eta$ for all 12 backbone/condition combinations on the Code2Video pipeline, across 2,252 total scene evaluations.

\begin{table}[h]
\centering
\caption{Scene-level completion ratio  $\eta = N_\text{ok}/N_\text{processed}$ across all  backbone/condition combinations. SGA achieves $\eta = 1.000$  for all four backbones; the two VLM Critic failures are unrelated to SGA's object extraction.}
\label{tab:eta}
\adjustbox{max width=\linewidth}{%
\setlength{\tabcolsep}{6pt}
\begin{tabular}{lccc}
\toprule
Backbone & RAW & VLM Critic & SGA \\
\midrule
GPT-5.1           & 144/144 = 1.000 & 143/143 = 1.000 & 288/288 = 1.000 \\
GPT-5 mini        & 151/151 = 1.000 & 151/151 = 1.000 & 303/303 = 1.000 \\
Gemini 3.0 Flash  & 121/121 = 1.000 & 121/121 = 1.000 & 242/242 = 1.000 \\
Claude 4.6 Sonnet & 147/147 = 1.000 & 145/147 = 0.986 & 294/294 = 1.000 \\
\midrule
\textbf{Total}
           & 563/563 = 1.000 & 560/562 = 0.996 & 1127/1127 = 1.000 \\
\bottomrule
\end{tabular}}
\end{table}

SGA achieves $\eta = 1.000$ across all four backbones (1,127 of 1,127 scenes completed), while the VLM Critic reaches $\eta = 0.996$ (2 failures in the Claude 4.6 Sonnet condition). Two properties of the pipeline design bound the risk of measurement bias regardless of $\eta$: (1) the scene graph is extracted from the \emph{original} LLM-generated code before patching (SGA only inserts \texttt{.move\_to()} calls, never removes objects) so any extraction incompleteness affects RAW and SGA identically; (2) omitting a misclassified object simultaneously penalizes both the overlap metric and the labeling completeness score, preventing systematic upward bias in MVQS.

% ============================================================
\section{Human Evaluation: Full Protocol and Detailed Breakdown}
\label{sec:supp_human}
% ============================================================

This appendix provides the complete evaluation protocol, participant details, inter-rater reliability metrics, and per-question splits for the human evaluation summarized in Section~\ref{sec:human_eval_main}.

\paragraph{Participant Pool and Setup.}
We recruited $25$ participants via a standardized web evaluation platform. A total of 16 participants completed all video-rating trials and 15 completed all frame-rating trials, with 14 participants ($56\%$) completing the entire five-page evaluation suite. Rather than discarding incomplete responses, we report results across all validated submissions to maximize statistical power without introducing selection bias.

\paragraph{Experimental Design.}
The evaluation is split into two distinct tracks:
\begin{enumerate}[leftmargin=*,itemsep=2pt,topsep=2pt]
    \item \textbf{Frame-Level Pairwise Comparisons ($n=45$ ratings per comparison):} Evaluators judged 8 pairwise frame comparisons ($16$ static frames total, disjoint from the video-rating set) targeting explicit geometric errors: \textit{Overlap} (element occlusions) and \textit{Canvas} (out-of-boundary errors). Raters made pairwise preference judgments or indicated a tie.
    \item \textbf{Video-Level Quality Ratings ($n=96$ ratings per comparison):} Evaluators assessed 6 randomly sampled video triplets ($\text{RAW}$, $\text{VLM Critic}$, $\text{SGA}$), yielding $18$ distinct clips. Each clip was rated on a 7-point Likert scale for \textit{Readability} and \textit{Pedagogical Clarity}. Raters also indicated their overall 3-way video preference.
\end{enumerate}
Two of the six video triplets ($t4$ and $t5$) intentionally contained valid, semantically necessary element overlaps to test whether each framework over-corrects non-faulty layouts.

\paragraph{Inter-Rater Reliability.}
We measure inter-rater agreement using Krippendorff's $\alpha$ (ordinal weighting, $95\%$ bootstrap confidence intervals with $1{,}000$ resamples) across all task types, detailed in Table~\ref{tab:supp_irr}. Frame-level judgments show fair-to-moderate consensus ($\alpha = 0.407$), driven primarily by high agreement on overlapping elements ($\alpha = 0.468$). Video-level metrics display weaker agreement ($\alpha = 0.208$), reflecting the higher subjective variance in evaluating multi-second animation clips versus static geometric defects.

\begin{table}[t]
\centering
\small
\setlength{\tabcolsep}{5pt}
\caption{Inter-rater reliability measured via Krippendorff's $\alpha$ (ordinal, 95\% bootstrap CI over 1,000 resamples).}
\label{tab:supp_irr}
\begin{tabular}{lccc}
\toprule
Evaluation Category & Raters & Units & $\alpha$ (95\% CI) \\
\midrule
\textbf{Frame Level (Pooled)} & 15 & 16 & \textbf{0.407}~[0.287, 0.606] \\
\quad Overlap & 15 & 8 & 0.468~[0.326, 0.697] \\
\quad Canvas & 15 & 8 & 0.353~[0.214, 0.566] \\
\midrule
\textbf{Video Level (Pooled)} & 16 & 36 & \textbf{0.208}~[0.091, 0.410] \\
\quad Readability & 16 & 18 & 0.248~[0.113, 0.456] \\
\quad Pedagogical Clarity & 16 & 18 & 0.161~[0.059, 0.368] \\
\bottomrule
\end{tabular}
\end{table}

\paragraph{Detailed Frame-Level Breakdown.}
Table~\ref{tab:supp_frame} decomposes the frame-level win rates reported in Table~\ref{tab:human_eval} of the main text. $\text{SGA}$ demonstrates its strongest advantage over $\text{RAW}$ on \textit{Overlap} resolutions ($93.3\%$), while achieving its largest margin over $\text{VLM Critic}$ on \textit{Canvas}-boundary violations ($73.3\%$). This reveals distinct failure modes between the baselines: while $\text{VLM Critic}$ occasionally resolves overlapping elements, its visual re-prompting frequently pushes elements beyond the canvas border, whereas $\text{SGA}$ enforces both constraints simultaneously.

\begin{table}[t]
\centering
\small
\setlength{\tabcolsep}{3pt} % Tightens inter-column spacing to prevent overflow
\caption{Detailed pairwise frame-level win, loss, and tie rates ($n=45$ ratings per comparison).}
\label{tab:supp_frame}
\begin{tabular}{@{} llccc @{}} % @{} removes padding at the outer edges
\toprule
Comparison & Error Type & Win (\%) & Loss (\%) & Tie (\%) \\
\midrule
\multirow{2}{*}{SGA vs.\ RAW} & Overlap & \textbf{93.3} & 2.2 & 4.4 \\
 & Canvas & \textbf{75.6} & 0.0 & 24.4 \\
\midrule
\multirow{2}{*}{SGA vs.\ VLM Critic} & Overlap & \textbf{56.7} & 13.3 & 30.0 \\
 & Canvas & \textbf{73.3} & 26.7 & 0.0 \\
\midrule
\multirow{2}{*}{RAW vs.\ VLM Critic} & Overlap & 26.7 & \textbf{46.7} & 26.7 \\
 & Canvas & 33.3 & \textbf{42.2} & 24.4 \\
\bottomrule
\end{tabular}
\end{table}

\paragraph{Video-Level Analysis and Over-Correction Case Study.}
Pairwise video-level head-to-head rates (pooled across Readability and Pedagogical Clarity, $n=192$ ratings per pair) are summarized in Table~\ref{tab:supp_video}. $\text{SGA}$ maintains a consistent preference margin over both baselines, though higher tie rates ($51.6\%$ vs.\ $\text{RAW}$) indicate that motion, timing, and narrative continuity dilute purely static spatial signals during clip-level observation.

To evaluate model resistance against false-positive edits, we isolate the $t4$ and $t5$ intentional-overlap trials (Table~\ref{tab:supp_t4t5}). Because $\text{SGA}$ evaluates symbolic scene graphs before applying refactoring, it correctly recognizes these intentional overlaps as semantically valid and leaves the layout intact, maintaining high scores (e.g., $6.27$ vs.\ $6.09$ on $t5$ Readability). Conversely, $\text{VLM Critic}$ misinterprets valid overlap as a visual error, triggering unnecessary layout modifications that degrade both Readability ($3.64$) and Pedagogical Clarity ($3.27$).

\begin{table}
\centering
\small
\setlength{\tabcolsep}{8pt}
\caption{Video-level paired head-to-head win rates, pooled across Readability and Pedagogical Clarity ($n=192$ ratings per comparison).}
\label{tab:supp_video}
\begin{tabular}{lccc}
\toprule
Comparison & Win (\%) & Loss (\%) & Tie (\%) \\
\midrule
SGA vs.\ RAW & \textbf{37.0} & 11.5 & 51.6 \\
SGA vs.\ VLM Critic & \textbf{55.2} & 17.7 & 27.1 \\
RAW vs.\ VLM Critic & \textbf{45.8} & 31.2 & 22.9 \\
\bottomrule
\end{tabular}
\end{table}

\begin{table}
\centering
\small
\setlength{\tabcolsep}{6pt}
\caption{Mean ratings (7-point Likert scale) on intentional-overlap case studies ($t4$, $t5$) testing over-correction resistance.}
\label{tab:supp_t4t5}
\begin{tabular}{llccc}
\toprule
Trial & Metric & RAW & SGA & VLM Critic \\
\midrule
\multirow{2}{*}{$t4$} & Readability & \textbf{5.91} & 5.82 & 3.82 \\
 & Pedagogical Clarity & 5.09 & \textbf{5.18} & 3.73 \\
\midrule
\multirow{2}{*}{$t5$} & Readability & 6.09 & \textbf{6.27} & 3.64 \\
 & Pedagogical Clarity & 5.82 & \textbf{6.09} & 3.27 \\
\bottomrule
\end{tabular}
\end{table}

% ============================================================
\section{Experimental Conditions and Baseline Fairness}
\label{sec:supp_fairness}
% ============================================================

Both SGA and the VLM Critic were evaluated under a fixed budget of $n=1$ refinement iteration applied identically across all pipeline/backbone combinations. For each topic, the procedure is:
(1) generate initial script $C^{(0)}$ with backbone $\mathcal{L}$;
(2) apply one round of feedback (SGA or VLM Critic);
(3) generate refined script $C^{(1)}$ with the same backbone $\mathcal{L}$;
(4) evaluate $C^{(1)}$ with both MVQS and VLM-Judge.

The SGA scene counts in Table~\ref{tab:fullscores_bench} are exactly $2\times$ the RAW counts for every backbone (e.g.,\ 288 vs.\ 144 for GPT-5.1), providing direct empirical confirmation that each topic was processed through exactly one generate--refine--evaluate loop.

Both methods consume exactly one LLM inference call per iteration. The only asymmetry is the rendering cost: the VLM Critic requires a full rendering pass ($\approx$30--90 seconds) before inference, whereas SGA uses partial execution ($<$5 seconds). This difference does not affect the number of refinement steps or the information available to the backbone during refinement; it affects only wall-clock latency, which is reported separately in Section~4.3 of the main paper.

\paragraph{VLM Critic Reproducibility.} 
To ensure baseline reproducibility, the VLM Critic is implemented using the native Code2Video Critic module invoked with its Visual Anchor Prompt. For each rendered section video, the model receives: (i)~a fixed grid reference overlay, (ii)~the section title, (iii)~the sequence of animation steps, and (iv)~the current grid-occupancy map. The critic outputs up to three structured layout corrections expressed as explicit grid coordinates. This structured feedback is then passed to the target LLM backbone $\mathcal{L}$ alongside the complete source code to complete the single refinement pass $C^{(1)}$.

\paragraph{MVQS Reproducibility.} 
MVQS is computed deterministically from the extracted scene graph $\mathcal{S}$ following Eq.~(1) and the aggregation weight vectors specified in Section~3.4 of the main text. Specifically, the evaluation utilizes spatial component weights $\mathbf{w}_\text{sp} = [0.50, 0.25, 0.15, 0.10]$, layout and edge component weights $\mathbf{w}_\text{ly} = \mathbf{w}_\text{ed} = [0.40, 0.60]$, and composite aggregation weights $\{0.5, 0.2, 0.3\}$. The scoring algorithm contains zero stochastic elements or sampling variations, guaranteeing that repeated execution over identical scene graphs yields exact score parity. The complete reference implementation, scoring metrics, and benchmark evaluation scripts are publicly accessible via our \href{https://jhedlp.github.io/projects/sga/}{project page}.

% ============================================================
\section{SGA Algorithm Pseudocode}
\label{sec:supp_algorithm}
% ============================================================

Algorithm~\ref{alg:sga} provides a self-contained pseudocode summary of the SGA pipeline described in Section~3 of the main paper.

\begin{algorithm}[h]
\caption{SGA pipeline (single iteration). If no violations are detected ($V^{(0)} = \emptyset$), the original script is returned unchanged, incurring no LLM refinement cost.}
\label{alg:sga}
\begin{algorithmic}[1]
\Require Topic $T$, generative model $\mathcal{L}$, conflict threshold $\varepsilon$
\Ensure Refined animation script $C^{(1)}$
\State $C^{(0)} \leftarrow \mathcal{L}(T)$ \Comment{Initial generation}
\State $\mathcal{S}^{(0)} \leftarrow \textsc{PartialExecute}(C^{(0)})$ \Comment{AST instrumentation + runtime extraction}
\State $V^{(0)} \leftarrow \textsc{DetectConflicts}(\mathcal{S}^{(0)},\,\varepsilon)$ \Comment{AABB overlap + semantic filter}
\If{$V^{(0)} = \emptyset$}
    \State \Return $C^{(0)}$ \Comment{No violations---skip refinement}
\EndIf
\State $F^{(0)} \leftarrow \textsc{CompileFeedback}(V^{(0)},\,\mathcal{S}^{(0)})$ \Comment{MSV vectors + fix templates}
\State $C^{(1)} \leftarrow \mathcal{L}\!\left(T,\,C^{(0)},\,F^{(0)}\right)$ \Comment{Targeted line-level refinement}
\State \Return $C^{(1)}$
\end{algorithmic}
\end{algorithm}

\paragraph{Notation.} 
$\textsc{PartialExecute}$ instruments the \texttt{construct()} method  with per-statement \texttt{try-except} blocks and extracts the symbolic  scene graph $\mathcal{S}^{(0)}$ via Manim runtime instrumentation, as  described in Section~3.2 of the main paper. 
$\textsc{DetectConflicts}$ applies exact AABB intersection analysis followed by the two-stage semantic and spatio-temporal filter  (Section~3.3).  
$\textsc{CompileFeedback}$ translates the violation set $V^{(0)}$ into  a structured diagnostic report containing pre-computed absolute target  coordinates $(t_x, t_y)$ and \texttt{fix\_target} identifiers  (Section~3.4). The SGA terminates after a fixed budget of $n$  iterations; in all experiments reported in the main paper, $n=1$.

\begin{table*}
\centering
\caption{Scene-level MVQS distributions (mean $\pm$ std) from  the full Code2Video benchmark (2,252 evaluations). SGA scene counts are $2\times$ RAW because each topic produces both a pre-patch and a post-patch evaluation.}
\label{tab:fullscores_bench}
\resizebox{\textwidth}{!}{%
\setlength{\tabcolsep}{6pt}
\begin{tabular}{ll c c cccc}
\toprule
Backbone & Condition & Scenes & $\eta$
  & MVQS & $S_\text{sp}$ & $S_\text{ly}$ & $S_\text{ed}$ \\
\midrule
\multirow{3}{*}{GPT-5.1}
  & RAW        & 144 & 1.000 & $62.9{\pm}9.8$  & $69.9{\pm}9.5$  & $65.2{\pm}10.6$ & $49.7{\pm}28.4$ \\
  & VLM Critic & 143 & 1.000 & $64.5{\pm}10.5$ & $71.3{\pm}9.9$  & $66.9{\pm}10.6$ & $51.4{\pm}29.6$ \\
  & SGA        & 288 & 1.000 & $73.0{\pm}16.3$ & $79.8{\pm}15.4$ & $68.0{\pm}8.9$  & $65.0{\pm}32.6$ \\
\midrule
\multirow{3}{*}{GPT-5 mini}
  & RAW        & 151 & 1.000 & $63.6{\pm}9.9$  & $73.1{\pm}11.2$ & $65.2{\pm}9.9$  & $46.7{\pm}25.3$ \\
  & VLM Critic & 151 & 1.000 & $63.5{\pm}10.2$ & $74.2{\pm}11.8$ & $66.3{\pm}9.8$  & $43.7{\pm}26.3$ \\
  & SGA        & 303 & 1.000 & $66.4{\pm}9.0$  & $76.0{\pm}10.7$ & $66.0{\pm}10.3$ & $50.8{\pm}22.7$ \\
\midrule
\multirow{3}{*}{Gemini 3.0 Flash}
  & RAW        & 121 & 1.000 & $67.8{\pm}12.4$ & $78.0{\pm}16.9$ & $64.2{\pm}10.5$ & $53.1{\pm}25.8$ \\
  & VLM Critic & 121 & 1.000 & $69.0{\pm}11.5$ & $78.8{\pm}16.2$ & $63.7{\pm}10.0$ & $56.1{\pm}27.0$ \\
  & SGA        & 242 & 1.000 & $71.8{\pm}11.4$ & $82.6{\pm}15.4$ & $64.0{\pm}10.1$ & $58.9{\pm}23.4$ \\
\midrule
\multirow{3}{*}{Claude 4.6 Sonnet}
  & RAW        & 147 & 1.000 & $63.2{\pm}10.1$ & $72.8{\pm}9.7$  & $68.2{\pm}8.2$  & $44.0{\pm}28.9$ \\
  & VLM Critic & 147 & 0.986 & $65.1{\pm}9.9$  & $74.3{\pm}10.4$ & $68.2{\pm}8.8$  & $47.8{\pm}29.0$ \\
  & SGA        & 294 & 1.000 & $65.8{\pm}10.8$ & $75.3{\pm}10.5$ & $69.2{\pm}9.8$  & $47.7{\pm}29.1$ \\
\bottomrule
\end{tabular}}
\end{table*}

% ============================================================
\section{Full Scene-Level Score Breakdown}
\label{sec:supp_fullscores}
% ============================================================

Table~\ref{tab:fullscores_bench} provides the scene-level MVQS distributions from the full Code2Video benchmark (2,252 evaluations), including per-backbone mean and standard deviation for all three conditions. These scene-level distributions are consistent with the per-configuration values reported in Table~1 of the main paper, confirming that the 24-configuration aggregate results generalize across the full benchmark distribution.

\paragraph{Note on scene counts.}
SGA scene counts are exactly $2\times$ the RAW counts for every backbone because each topic produces both a pre-patch evaluation of $C^{(0)}$ and a post-patch evaluation of $C^{(1)}$. This design allows direct measurement of SGA's per-scene improvement and is consistent with the single-iteration protocol detailed in Section~\ref{sec:supp_fairness}.